\documentclass{ieeetj}
\usepackage[T1]{fontenc}
\usepackage[utf8]{inputenc}
\usepackage{microtype}
\usepackage{graphicx}
\usepackage{placeins}
\usepackage{booktabs}
\usepackage{stfloats}
\usepackage{amsmath,amssymb}
\usepackage{siunitx}
\usepackage{cite}
\usepackage[dvipsnames]{xcolor}
\usepackage{hyperref}
\usepackage{balance}
\usepackage{etoolbox}
\apptocmd{\thebibliography}{\interlinepenalty500\relax}{}{}

\hypersetup{
  hidelinks,
  pdfauthor={},
  pdftitle={An Open Panoramic Aerial Robot},
  pdfsubject={IEEE Transactions on Field Robotics submission},
  pdfkeywords={field robotics, panoramic perception, unmanned aerial vehicles}
}

\def\OJlogo{}
\def\seclogo{}
\long\def\receivedfont#1\par{}

\makeatletter
\def\ps@review{%
  \def\@oddhead{}\let\@evenhead\@oddhead
  \def\@oddfoot{\hfil\normalfont\small\thepage\hfil}%
  \let\@evenfoot\@oddfoot}
\let\ps@plain\ps@review
\def\p@subsection{\thesection-}
\makeatother
\begin{document}

\title{An Open Panoramic Aerial Robot: Airframe-Integrated Multi-Fisheye
Sensing, Onboard ERP Formation, and Field Evaluation}

\author{Dun Dai$^{1,*}$, Ze Lu$^{1,*}$, Cheng He$^{2}$, Yaowen Wang$^{1}$, and Quan Quan$^{1,2,\dagger}$}

\affil{School of Automation Science and Electrical Engineering, Beihang University, Beijing 100191, China}

\affil{Tianmushan Laboratory, Hangzhou 311115, China}

\authornote{$^{*}$Dun Dai and Ze Lu contributed equally to this work.}

\corresp{$^{\dagger}$Corresponding author: Quan Quan (e-mail: qq\_buaa@buaa.edu.cn).}

\begin{abstract}
We present an open panoramic aerial robot with four synchronized fisheye
cameras integrated into a carbon-fiber airframe and an onboard NVIDIA Jetson Orin NX.
The robot outputs calibrated raw views and an equirectangular panorama (ERP;
$1280\times640$ in all experiments). The ERP pipeline uses overlap-specific projection radii, gated
local alignment, seam control, and multi-rate state updates, and it runs
onboard on the live four-camera stream during flight. The field dataset
contains 18 sequences and more than 50,000 synchronized groups from seven
sites. On a 60-frame far-field sample, the method reduces the median per-frame
AKAZE P90 misalignment by 40.7\% compared with Fixed Radius, and with fixed
parameters it gives the lowest geometric errors among the tested controls at
two held-out sites. Controlled replay on the same NVIDIA Jetson Orin NX measures final-ERP
continuity, timing, and module-input power at a \SI{20}{\hertz} input rate.
Compared with external stitching software given the same calibrated
projection, the onboard pipeline gives final-ERP line continuity no lower than
any tested method, while every external configuration measured on the module
needs 5.3 to 147 times the input period and 3.8 to 106 times the energy per
output. Frozen detection and place-recognition models are used to evaluate the
exported images. Code, calibration, reference hardware, and data-access
documentation are available in an anonymized repository at
\url{https://anonymous.4open.science/r/Open-Pano-Field-CE1F/README.md}.
\end{abstract}


\maketitle
\section{Introduction}

Ultra-low-altitude unmanned aerial vehicles (UAVs) fly close to buildings,
vegetation, and people, and a wide field of view keeps visual context when the
vehicle turns \cite{lin2018aerialnavigation,petrlik2025subterranean}. For a
panoramic camera rig, camera placement decides overlap and parallax,
synchronization and calibration decide how well images can be aligned, and
the onboard computer decides output rate and power. The hardware and the
image processing therefore have to be designed together.

\begin{figure}[!t]
  \centering
  \includegraphics[width=\columnwidth]{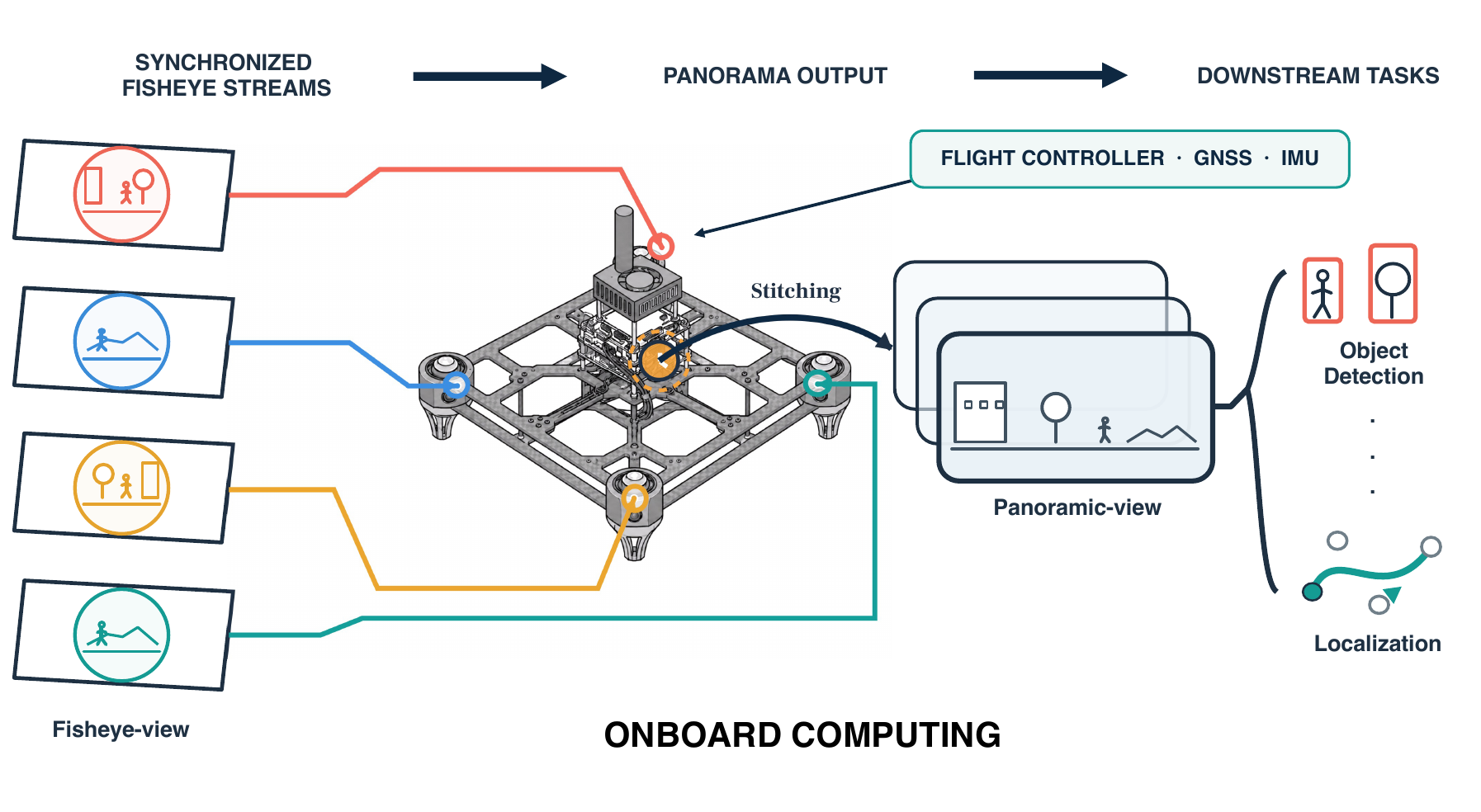}
  \caption{Open panoramic aerial robot. Synchronized fisheye views and calibration
  support camera-level processing; the ERP interface supports detection and
  place recognition.}
  \label{fig:teaser}
\end{figure}

Raw or rectified views suit algorithms that model each camera separately. An
ERP gives a shared angular grid and can be resampled into perspective
sectors. Our robot provides both interfaces (Fig.~\ref{fig:teaser}).

Existing aerial panoramic systems have been used for segmentation and
visual--inertial simultaneous localization and mapping (SLAM)
\cite{wang2022aerialpass,wu2026panoair}. To reuse such a system, one also
needs its camera geometry, its image formation code, and its measured
operating cost. This work provides these together: synchronized sensing,
onboard panorama formation, and field evaluation.

The main contributions are:
\begin{itemize}
  \item A physical four-fisheye aerial platform with live onboard ERP
  formation, synchronized raw-view and ERP outputs, calibration, source
  timestamps, and released software and reference CAD. Its sensing layout and
  ERP geometry follow the public RflyPano simulation benchmark
  \cite{dai2026rflypano}, so methods developed on that benchmark can be
  applied to this platform's field data.
  \item A rig-aware ERP formation pipeline with overlap-specific projection
  radii, gated local correction, and seam-controlled fusion, implemented on the
  onboard NVIDIA Jetson Orin NX with cached maps and multi-rate state updates.
  \item A field evaluation with live onboard operation, image-quality
  comparisons on common inputs, temporal-policy tests, repeated NX
  measurements of final-ERP quality, latency, and energy, a comparison with
  external stitching software at a common composition interface, and
  frozen-model tests of the exported images.
\end{itemize}

\section{Related Work}

\subsection{Panoramic sensing and formation for field robots}

Aerial datasets often combine perspective cameras with inertial, GNSS, or
LiDAR sensors \cite{li2024marslvig,wang2025uavscenes}. Aerial-PASS uses a
panoramic annular lens for segmentation, and PAIR360 provides six fisheye
views on a ground platform \cite{wang2022aerialpass,kim2024pair360}.
FishEye8K supports multi-fisheye perception \cite{gochoo2023fisheye8k}.
PanoAir uses a dual-fisheye camera for embedded panoramic visual--inertial
SLAM \cite{wu2026panoair}. OmniNxt is an open multi-fisheye aerial platform
that outputs fisheye images only and forms no ERP; its odometry and dense
mapping use virtual views derived internally
\cite{liu2024omninxt,omninxt2026hardware}. Consumer 360-degree cameras such as
Insta360 X5 and DJI Osmo 360 \cite{insta360x5_2025,djiosmo360_2025} output
stitched panoramas, but they are closed products that expose neither
synchronized raw fisheye views nor calibration, and they are not integrated
with an airframe or an onboard computer.

Other field robots put wide-view sensing into complete navigation systems.
Lin et al. combine fisheye imaging, state estimation, dense mapping, and
planning on a quadrotor \cite{lin2018aerialnavigation}, and Petrl{\'i}k et al.
use four RGB streams in a subterranean multi-UAV system
\cite{petrlik2025subterranean}. Our focus is narrower: an accessible
sensing-to-ERP system, its image quality, and its embedded cost.
Table~\ref{tab:related-system-comparison} compares the reported interfaces.

\begin{table}[!b]
  \centering
  \caption{Reported availability and output interfaces. \checkmark: reported
  available; --: not established in the cited source. Real data means images
  captured in the physical world, not rendered. For Ours, HW is the reference
  hardware CAD and data is a separately hosted dataset.}
  \label{tab:related-system-comparison}
  \footnotesize
  \setlength{\tabcolsep}{0pt}
  \begin{tabular*}{\columnwidth}{@{\extracolsep{\fill}}lcccccc@{}}
    \toprule
    System/product & \shortstack{Real\\data} & \shortstack{HW\\open} & \shortstack{Code\\open} & \shortstack{Data\\open} & \shortstack{ERP\\out.} & \shortstack{Fisheye\\out.} \\
    \midrule
    PanoAir \cite{wu2026panoair} & \checkmark & -- & \checkmark & \checkmark & \checkmark & \checkmark \\
    RflyPano \cite{dai2026rflypano} & -- & -- & \checkmark & \checkmark & \checkmark & \checkmark \\
    OmniNxt \cite{liu2024omninxt} & \checkmark & \checkmark & \checkmark & -- & -- & \checkmark \\
    Omni-Swarm \cite{xu2022omniswarm} & \checkmark & -- & \checkmark & \checkmark & -- & \checkmark \\
    Insta360 X5/DJI Osmo 360 \cite{insta360x5_2025,djiosmo360_2025} & \checkmark & -- & -- & -- & \checkmark & -- \\
    \textbf{Ours} & \checkmark & \checkmark & \checkmark & \checkmark & \checkmark & \checkmark \\
    \bottomrule
  \end{tabular*}
\end{table}

Parallax-aware stitching uses spatially varying warps, structure
constraints, and learned alignment \cite{zaragoza2013apap,lin2015aanap,
du2022gesgsp,nie2023parallaxstitching,kweon2023pixelwarping}, and patch
alignment, gravity priors, and learned matching add further geometric
constraints \cite{liao2025patchseam,ding2021gravitystitching,
lindenberger2023lightglue}. These methods estimate the warp from image
content. On a calibrated rig this warp is known in closed form, so a fair
comparison on our platform is at the composition stage, where all methods get
the same projected layers. Our pipeline combines calibrated overlap-wise
projection, gated local correction, and scheduled state updates for
continuous four-camera input. Sec.~\ref{sec:experiments} isolates
projection, seam, and local-correction choices inside one calibrated
pipeline, and Sec.~\ref{sec:external-baselines} compares with external
methods at the composition stage, including a recent seam-cutting method
\cite{liao2025patchseam}.

\subsection{Embedded panoramic interfaces for robot perception}

ERP perception methods handle projection distortion with adaptation,
spherical geometry, and attention
\cite{zhang2022trans4pass,li2023sgat4pass,zhang2024goodsam,benny2025sphereuformer}.
Panoramic depth estimation also uses projection and cross-view constraints
\cite{reyarea2022_360monodepth,cao2025panda}. Perspective descriptors and
panoramic retrieval benchmarks support place-recognition research
\cite{alibey2023mixvpr,keetha2024anyloc,huang2024_360loc}, and field studies
evaluate panoramic localization and long-term robot operation
\cite{grelsson2020horizonnet,mattamala2025forest}. AirSim360 is an aerial
panoramic simulation platform \cite{ge2026airsim360};
Sec.~\ref{sec:rflypano-configuration} discusses the simulated counterpart of
our sensing setup. In this work, frozen detection and retrieval models test
the image interface of the physical robot, together with direct measurements
of formation quality and runtime.

\subsection{Relation to simulation benchmarks and open platforms}
\label{sec:rflypano-configuration}

RflyPano \cite{dai2026rflypano} is a simulation benchmark for
ultra-low-altitude UAV localization rendered in RflySim. It provides synthetic
four-view fisheye sequences, a $1280\times640$ ERP, reference poses, and a
camera model that maps ERP directions through camera rotations and polynomial
fisheye projection. It has no physical airframe, hardware timing, real rig
calibration, or field capture. Our work is a physical platform: a carbon-fiber
quadrotor with four hardware-synchronized fisheye cameras, onboard ERP
formation in flight, and a seven-site field dataset. No image, calibration, or
trajectory in this paper comes from the benchmark; all data were captured on
the physical robot.

\begin{figure}[!t]
  \centering
  \includegraphics[width=\columnwidth]{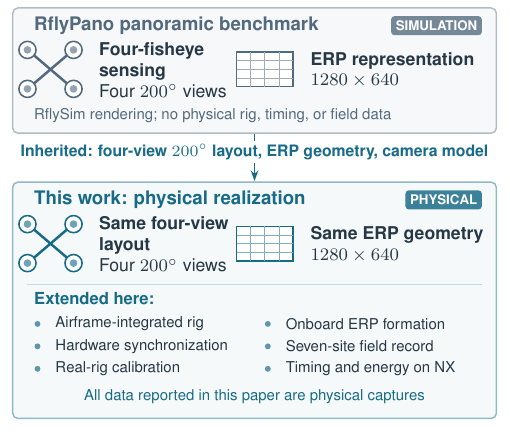}
  \caption{Simulation benchmark and physical platform. RflyPano
  \cite{dai2026rflypano} renders four $200^{\circ}$ fisheye views and a
  $1280\times640$ ERP in simulation. Our platform inherits this layout, ERP
  geometry, and camera model and extends them to a physical airframe with
  hardware synchronization, rig calibration, onboard formation, and field
  capture. All data in this paper are from the physical robot.}
  \label{fig:rflypano-configuration}
\end{figure}

The platform builds on this benchmark. It inherits the four-view
$200^{\circ}$ layout, the ERP geometry, and the camera model, and realizes
them on a physical airframe (Fig.~\ref{fig:rflypano-configuration}), so
methods developed on RflyPano can be applied to the field data of this robot.
The airframe integration, hardware synchronization, rig calibration,
overlap-wise projection, local correction, seam-controlled fusion, and
embedded scheduling are contributed here.

Among physical platforms, the closest is OmniNxt
\cite{liu2024omninxt,omninxt2026hardware}, an open quadrotor with four fisheye
cameras and released hardware and software. Its output is the fisheye images
only: virtual pinhole views are formed internally for odometry and dense
mapping, and the cited sources report neither an ERP output nor a released
field dataset. Our platform instead forms a calibrated
ERP onboard, exports it with the raw views and their source timestamps, and
releases a field dataset from seven sites under day and night conditions.

\section{System Design and Field Dataset}
\label{sec:platform-dataset}

\begin{figure*}[t]
  \centering
  \includegraphics[width=\textwidth]{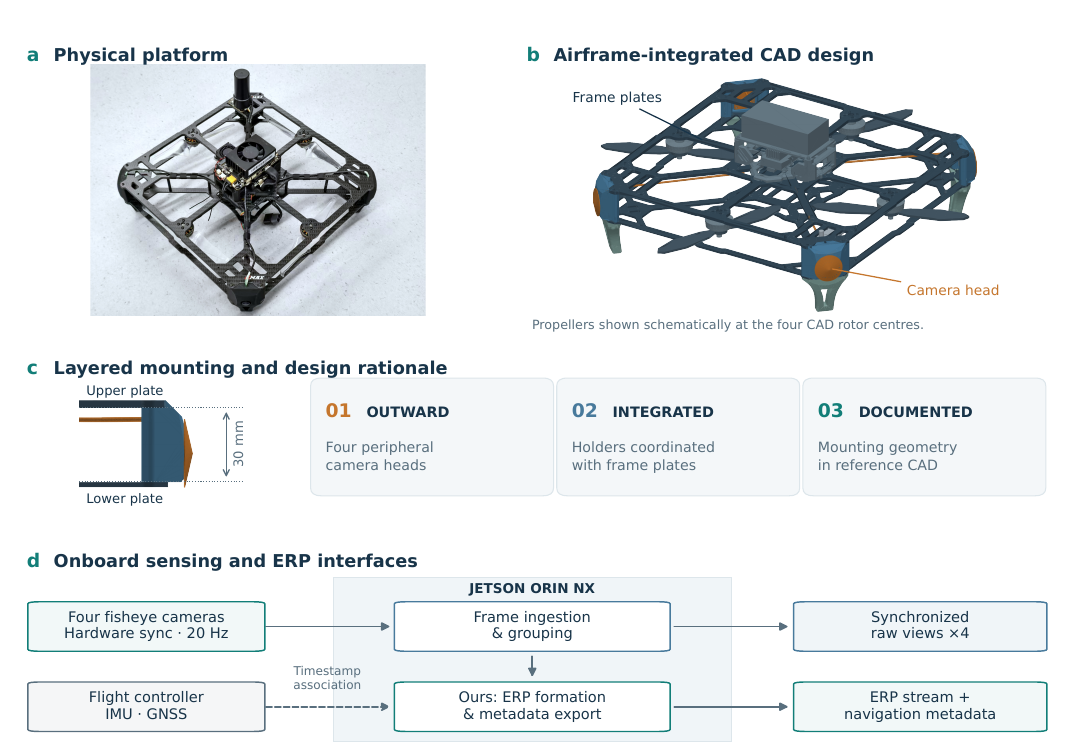}
\caption{Panoramic aerial robot. (a) Physical platform. (b) Reference CAD;
  propellers are shown schematically at the four rotor centres. (c) Camera holder and frame plates,
  with 30\,mm between facing plate surfaces. (d) Onboard image and metadata
  interfaces. The ERP stream was produced onboard during field operation;
  navigation records are associated by timestamp.}
  \label{fig:platform}
\end{figure*}

The robot combines synchronized imaging, onboard ERP formation, and
timestamped navigation records (Fig.~\ref{fig:platform}), and outputs raw
fisheye views, calibration, and ERP images.

\subsection{Integrated sensing, computing, and interfaces}
\label{sec:integrated}

The carbon-fiber frame holds four camera heads facing front-left,
front-right, rear-right, and rear-left. The robot is
$28\times28\times13.3$\,cm and weighs \SI{1002.3}{\gram} with its 4000 mAh
battery. It carries up to \SI{500}{\gram} of extra payload, and its endurance
is \SI{9}{\minute} without extra payload with the NVIDIA Jetson Orin NX in \SI{25}{\watt}
mode (Table~\ref{tab:platform-specification}). The cameras give
hardware-synchronized global-shutter images of $1088\times1280$ at
\SI{20}{\hertz} over USB~3.0.

Each camera points between two adjacent motor directions. The camera holders
are fixed to the upper and lower frame plates, whose facing surfaces are
30\,mm apart (Fig.~\ref{fig:platform}(b)--(c)), and the released CAD contains
these parts and their relations.

In flight, the 16-GB NVIDIA Jetson Orin NX receives the live synchronized
four-camera groups and runs the ERP pipeline onboard, targeting one
$1280\times640$ ERP per \SI{20}{\hertz} input group with a bounded queue.
Sec.~\ref{sec:embedded} describes the scheduling, and
Sec.~\ref{sec:nx-evaluation} reports the controlled replay measurements. A
MicoAir743v2-AIO-45A flight controller and an MG-F10-A GNSS receiver provide
navigation data \cite{micoair2026h743,micoair2026mgf10}. The software release
contains the stitching entry point, calibration, and alignment and colour
settings under the MIT licence. The hardware release contains an AP242 STEP
assembly, a checksum, a component list, and safety notes; custom hardware uses
CERN-OHL-S v2.0, and third-party components keep their own rights. Field data
are shared through a separate access route with a dataset notice.

\begin{table}[t]
  \centering
  \caption{Integrated platform configuration.}
  \label{tab:platform-specification}
  \footnotesize
  \begin{tabular}{@{}p{0.29\columnwidth}p{0.65\columnwidth}@{}}
    \toprule
    Component & Configuration \\
    \midrule
    UAV platform & Purpose-built carbon-fiber quadrotor frame with
    four embedded camera heads; $28\times28\times13.3$\,cm;
    \SI{1002.3}{\gram} with battery; \SI{693.2}{\gram} without battery \\
    Payload/propulsion & Up to \SI{500}{\gram} auxiliary payload;
    thrust-to-weight ratio 4.2:1 \\
    Unloaded endurance & \SI{9}{\minute} with no auxiliary payload and the
    NVIDIA Jetson Orin NX configured in \SI{25}{\watt} mode \\
    Panoramic camera & Four hardware-synchronized global-shutter fisheye views \\
    Visual input & $4\times1088\times1280$ at \SI{20}{\hertz};
    nominal \SI{200}{\degree} FOV; USB~3.0 \\
    Onboard computer & NVIDIA Jetson Orin NX Developer Kit; 16\,GB memory;
    configured \SI{25}{\watt} power mode \\
    Flight/navigation & MicoAir743v2-AIO-45A flight controller and MG-F10-A GNSS receiver \\
    Panorama output & $1280\times640$ at \SI{20}{\hertz} target \\
    Runtime environment & Ubuntu 20.04 (L4T 35.5); CUDA 11.4; TensorRT 8.5.2.2 \\
    \bottomrule
  \end{tabular}
\end{table}

\subsection{Synchronization and calibration}
\label{sec:synchronization}

Each synchronized four-view group has an image timestamp $t_g$. Navigation
data use this clock directly or an affine map
$\tilde t_k^{\rm nav}=a t_k^{\rm nav}+b$, where drift $a$ and offset $b$ are
anchored by shared changes of gyroscope magnitude. For
$\tilde t_k^{\rm nav}\leq t_g\leq\tilde t_{k+1}^{\rm nav}$, quaternion signs
are chosen so that $\mathbf q_k^\top\mathbf q_{k+1}\geq0$, and position and
attitude are interpolated as
\begin{equation}
 \begin{aligned}
 \lambda&=\frac{t_g-\tilde t_k^{\rm nav}}
 {\tilde t_{k+1}^{\rm nav}-\tilde t_k^{\rm nav}},\\
 \mathbf p_g&=(1-\lambda)\mathbf p_k+\lambda\mathbf p_{k+1},\\
 \Omega&=\arccos(\mathbf q_k^\top\mathbf q_{k+1}),\\
 \mathbf q_g&=\operatorname{SLERP}(\mathbf q_k,\mathbf q_{k+1};\lambda)\\
 &=\frac{\sin((1-\lambda)\Omega)}{\sin\Omega}\mathbf q_k
 +\frac{\sin(\lambda\Omega)}{\sin\Omega}\mathbf q_{k+1}.
 \end{aligned}
 \label{eq:pose-interpolation}
\end{equation}
where $\mathbf p_k$ and $\mathbf q_k$ are position and unit attitude
quaternion. The sign choice gives shortest-arc spherical linear interpolation
(SLERP); when $\Omega\to0$, normalized linear interpolation is used as the
limit. Associations outside the recorded interval or across gaps longer than
\SI{0.2}{\second} are rejected, and the original timestamps and association
status are kept.

Kalibr gives the camera models and extrinsics \cite{furgale2013calibration}.
Camera order, intrinsics, extrinsics, and image-space refinement are fixed in
each evaluation, and all transforms use the left-front camera $C_0$ as
reference. The image-quality and runtime tests use only visual input.

\subsection{Field data for evaluation}
\label{sec:field-data}

The field dataset contains 18 sequences and more than 50,000 synchronized
four-view groups from seven sites: Court, Farmland, Lake, North Playground,
North Square, South Playground, and South Square
(Fig.~\ref{fig:dataset-overview}). South Playground and South Square were
flown both at day and at night.

\begin{figure*}[t]
  \centering
  \includegraphics[width=\textwidth]{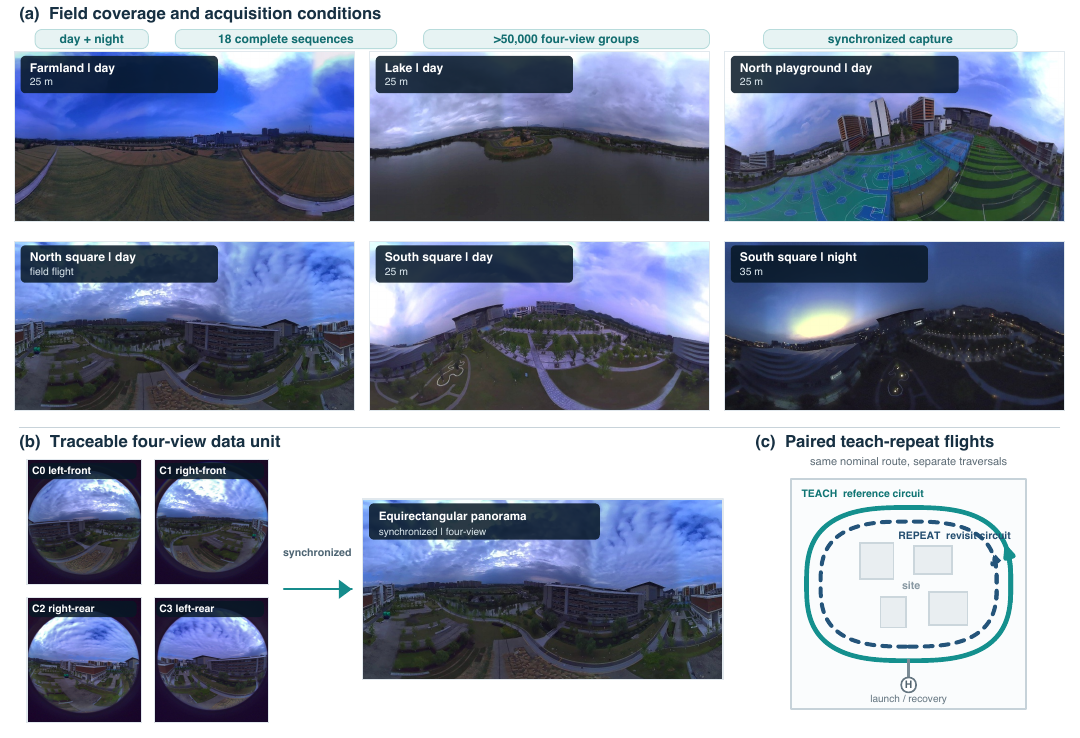}
  \input{figures/final/Fig3_Dataset_Overview_caption.tex}
\end{figure*}

The dataset includes raw images, panoramas, calibration, and navigation data
where available, and subsets are defined by sequence or site. Farmland is used
for development, and Lake and South Square are held out for quality tests.
Sec.~\ref{sec:experiments} gives the subsets used in each comparison.

\section{Onboard Multi-Fisheye Panorama Formation}
\label{sec:adaptive-panorama}

The pipeline maps each calibrated four-fisheye group to an ERP and keeps the
raw views. It uses a calibrated projection grid and an overlap-wise radius
bank; Sec.~\ref{sec:embedded} describes caching and scheduling.

Let $I_i^t$ be the image of camera $C_i$ at time $t$, $i\in\{0,1,2,3\}$. The
reference frame $\mathcal B$ has its origin at the optical centre of $C_0$,
and the output is $P^t\in\mathbb{R}^{H\times W\times3}$ with $H=640$ and
$W=1280$. A far projection works for distant content, but near objects need
local parallax correction. We therefore select one radius for each adjacent
overlap, apply it only in a narrow band, and keep the far projection
elsewhere.

\begin{figure*}[t]
  \centering
  \includegraphics[width=\textwidth]{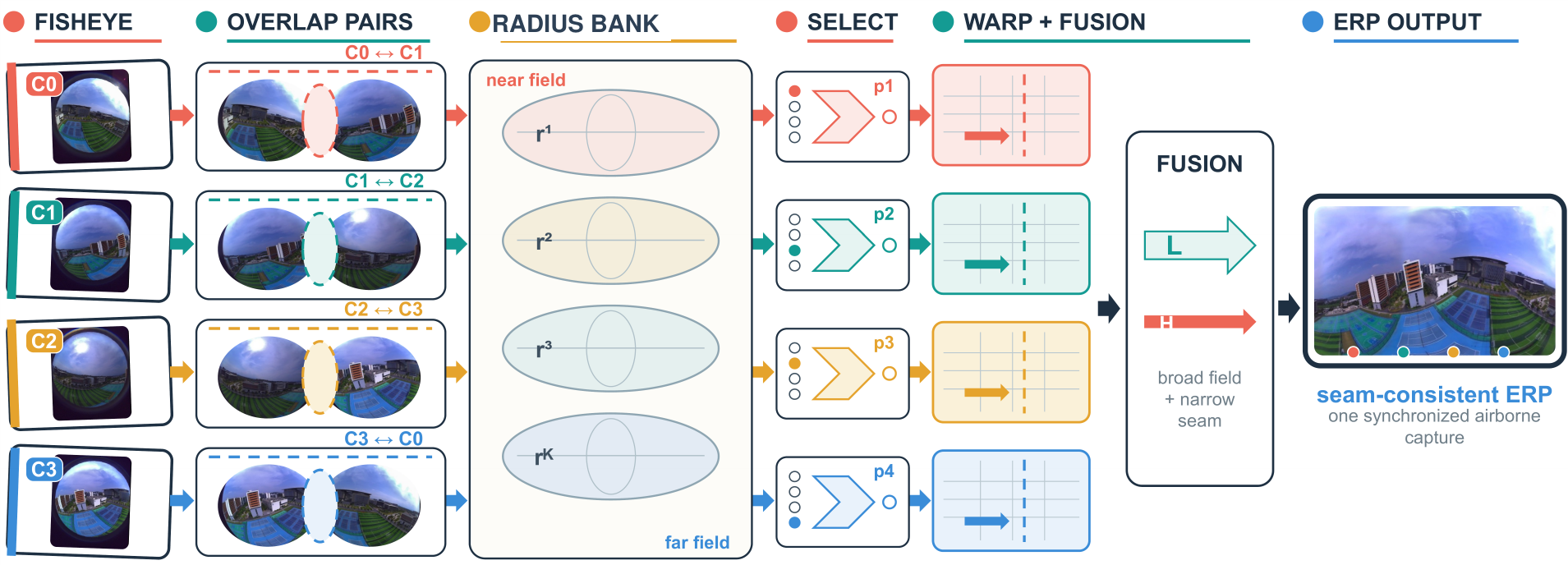}
\caption{Panorama formation pipeline. Each overlap selects a radius
  from the bank. Gated geometry updates, exposure correction, seam selection,
  and two-band fusion produce the ERP. Images are from an airborne capture;
  the rig and radius bank are schematic.}
  \label{fig:adaptive-stitching}
\end{figure*}

Figure~\ref{fig:adaptive-stitching} shows the pipeline. Feature matches
select the radii, with a photometric fallback and temporal confirmation.
Local mesh correction, exposure compensation, seam selection, and two-band
fusion then form the output. Accepted states are cached between updates.

\subsection{Rig-aware projection and adaptive radius}
\label{sec:radius}

For ERP pixel $(u,v)$, the longitude is $\theta=2\pi u/W-\pi$ and the
elevation is $\phi=\Phi(v/(H-1)-1/2)$, which give the unit ray in
$\mathcal B$
\begin{equation}
 \mathbf d(u,v)=\begin{bmatrix}\cos\phi\sin\theta&\ \sin\phi&\
 \cos\phi\cos\theta\end{bmatrix}^{\mathsf T}.
 \label{eq:erp-ray}
\end{equation}
where $\Phi$ is the vertical span. The radius $r$ is a discrete projection
hypothesis that approximates the overlap geometry, and the selected radius
sets the local warp inside the overlap band. The calibrated transform
$\mathbf T_{C_i\leftarrow\mathcal B}=[\mathbf R_i,\mathbf t_i]$ and the
equidistant camera model project $r\mathbf d$ into camera $C_i$. Remapping
tables and masks are cached for $\mathcal R=\{1,2,4,12,30\}\,$m.

For each adjacent-camera overlap $p$, bidirectional SIFT matches
\cite{lowe2004sift} are undistorted into rays. A ray $\mathbf d_{i,n}$ from
camera centre $\mathbf c_i$ hits the sphere of radius $r$ at
\begin{equation}
 \begin{aligned}
 \mathbf x_{i,n}(r)&=\mathbf c_i+\tau_{i,n}(r)\mathbf d_{i,n},\\
 \tau_{i,n}(r)&=-\mathbf c_i^{\mathsf T}\mathbf d_{i,n}+
 \sqrt{(\mathbf c_i^{\mathsf T}\mathbf d_{i,n})^2+r^2-
 \lVert\mathbf c_i\rVert_2^2}.
 \end{aligned}
 \label{eq:ray-sphere}
\end{equation}
After ERP projection $\pi(\cdot)$, let $e_{p,n}(r)$ be the Euclidean
displacement with horizontal wrap. Each overlap selects
\begin{equation}
 S_p(r)=Q_{0.65}\{e_{p,n}(r)\},\qquad
 \hat r_p=\arg\min_{r\in\mathcal R}S_p(r).
 \label{eq:radius-score}
\end{equation}
where $Q_{0.65}$ is the 65th percentile of valid ERP match displacements in
pixels. Matches that fit no layer are rejected. A candidate needs at least
eight valid matches and an improvement of 0.18 pixels. With
$\Delta S_p=S_p(r_p^{t-1})-S_p(\hat r_p)$, a feature-supported candidate is
accepted after one update if $\Delta S_p\geq0.5$ pixels, and otherwise after
two consecutive selections. Radius estimation runs every five frames. When
features are insufficient, a photometric fallback scores each radius by the
median seam cost (\ref{eq:seam-image-cost}) over its valid overlap and always
needs two consecutive confirmations (Supplementary Sec.~S2).
Sec.~\ref{sec:experiments} compares the temporal acceptance rules.

All thresholds in this section were set during development on the Farmland
sequences and then frozen. The main text keeps the values that define the
method's behaviour; Supplementary Secs.~S1--S2 list the complete settings.
The held-out Lake and South Square clips (Sec.~\ref{sec:stitching-comparison})
test whether these settings transfer to unseen sites.

\subsection{Seam-controlled photometric fusion}
\label{sec:fusion}

The \SI{30}{\metre} layer forms the base panorama. Inside a 72-pixel
half-band, the selected radius replaces the base warp with a 12-pixel
raised-cosine transition. Camera gains are estimated from robust
log-intensity differences between overlaps, combined with static colour
calibration, and updated every ten frames with a log-domain moving average
(Supplementary Sec.~S2). The seam cost is computed on the corrected overlaps.

Let $\delta_W$ be the signed circular column displacement on ERP width $W$,
and $B_s$ the seam-search half-width. The normalized centre offset is
$d_{c,y}=\delta_W(s_y^t,c_p)/B_s$ and the temporal offset is
$d_{t,y}=\delta_W(s_y^t,s_y^{t-1})/(2B_s)$. Dynamic programming minimizes
\begin{equation}
 \hat{\mathbf s}^{\,t}=\arg\min_{\mathbf s^t}\sum_y
 \left[C_p^t(y,s_y^t)+\lambda_c d_{c,y}^2+\lambda_t d_{t,y}^2\right],
 \label{eq:seam-objective}
\end{equation}
where $\mathbf s^t=\{s_y^t\}_y$ is the path and $c_p$ is the overlap centre.
Let $D_c$ be the mean absolute RGB difference divided by 255,
$D_s=G_4\ast D_c$, and $D_f=\mathcal D_{18,6}(D_c)$. With the normalized
gradient difference $D_g$ and edge strength $E$, the cost is
\begin{equation}
 C_p^t(y,x)=D_c+0.75D_s+0.35D_f+0.30D_g+0.35E.
 \label{eq:seam-image-cost}
\end{equation}
Here $G_4$ is Gaussian smoothing with $\sigma=4$ pixels and
$\mathcal D_{18,6}$ is a max filter with horizontal/vertical radii of 18/6
pixels. $D_g$ and $E$ are the normalized difference and maximum of the two
Sobel gradient magnitudes (Supplementary Sec.~S2), so all terms are
dimensionless. We use $\lambda_c=0.10$ and $\lambda_t=0.20$.
Per-seam uses $B_s=72$ pixels; the adaptive seam search in Ours tests 72 and
120 pixels and takes the wider path if its mean robust cost is at least 0.01
lower. Adjacent rows differ by at most 2 pixels, and the path moves at most
16 pixels between frames.

\paragraph{Gated local mesh.}
Ours adds a residual-triggered local correction before adaptive seam
selection. It is triggered only when at least 15 mutual controls remain and
their median displacement is above 8 pixels. The controls are split spatially
into about two thirds for fitting and one third for held-out validation;
distance-weighted local affine maps fitted on a grid give a dense overlap
flow. A pairwise mesh is accepted only if it reduces the validation median
error by at least 1 pixel and 12\%, and does not raise the P90 error by more
than 2\%. The accepted flow is applied inside the overlap band, and adaptive
seam selection is rerun; the mesh is kept only if the selected-path cost is
at most 0.005 higher than without it, otherwise the uncorrected overlap is
restored. The resulting geometry is evaluated with separate AKAZE features.
Supplementary Sec.~S1 gives the objective, all parameters, and pseudocode.

The corrected warp is split into
$L_i=G_\sigma\ast(M_iI_i)/(G_\sigma\ast M_i)$ and $H_i=I_i-L_i$, where $M_i$
is the validity mask and $G_\sigma$ is a mask-normalized Gaussian low-pass
with $\sigma=8$ pixels and horizontal wrap; division is pointwise where the
convolved mask is nonzero. The output is
\begin{equation}
 P^t=\sum_i w_i^{\mathrm{wide}}L_i+
     \sum_i w_i^{\mathrm{narrow}}H_i,
 \label{eq:two-band-blend}
\end{equation}
with 16- and 3-pixel wrapped feathers for the low and high bands. The
weights are zero outside valid masks and sum to one at each covered pixel,
which smooths colour transitions and keeps a narrow blend at edges.

\subsection{Panorama and navigation interfaces}

Raw views and ERP outputs keep the source image timestamp. Flight-controller
and GNSS records are associated with image groups as in
Sec.~\ref{sec:synchronization} and exported as timestamped navigation
metadata.

\subsection{Embedded implementation}
\label{sec:embedded}

The CUDA implementation separates per-group rendering from scheduled state
estimation. Rig maps, masks, and buffers are cached. Each group is uploaded
and remapped as an FP16 batch with the five-layer projection bank, and the
renderer combines the \SI{30}{\metre} base layer with the selected overlap
bands using the cached exposure and seam states.

The processed-group index increases by one for each group taken from the
input queue. SIFT jobs are submitted asynchronously every five processed
groups, nominally every \SI{250}{\milli\second} at \SI{20}{\hertz}. The first
group is initialized synchronously; later groups submit a job only when no
earlier job is still running. A finished job returns candidates tagged with
its source-group index, and on a later group the hysteresis of
Sec.~\ref{sec:radius} may accept a candidate. That group is rendered with the
new state, or with the cached state if nothing is accepted. Exposure gains
are updated every ten groups (nominally \SI{500}{\milli\second},
Sec.~\ref{sec:fusion}). Frame counting, job submission, job completion, state
acceptance, and rendering are therefore separate events, and each group is
rendered from its own images with the latest accepted states.

The input queue holds two groups; when it is full, the oldest group is
dropped. The drop rate is
$(N_{\mathrm{input}}-N_{\mathrm{published}})/N_{\mathrm{input}}$, and a late
output is counted separately from a dropped input. In the controlled replay,
sensor-rate operation means an ERP output rate within 0.1\,Hz of the
\SI{20}{\hertz} input, with the bounded queue and the reported drop rate.
Sec.~\ref{sec:nx-evaluation} gives the replay protocol and timing boundary.

\section{System and Experimental Evaluation}
\label{sec:experiments}

The evaluation links live onboard operation with controlled tests on
recorded flight data. In flight, the NVIDIA Jetson Orin NX forms ERPs from the live camera
stream. The recorded groups give identical inputs for image-quality
comparisons and for controlled \SI{20}{\hertz} replay on the NX, which
measures final-ERP quality, latency, throughput, and module-input power under
matched conditions. Each test fixes its inputs, calibration, and output
sampling, and detection and retrieval models are frozen.

The data unit is one synchronized four-view group.
Table~\ref{tab:stitching-summary} reports geometry on 60 far-field and 58
near-field common valid frames, detection on 181 matched frames, and angular
metrics on a 33-frame subset. Table~\ref{tab:cross-site-quality} covers the
held-out Lake and South Square clips. Figure~\ref{fig:task-level-benefits}
pools structure counts over a separate continuous near-field sequence, and
Table~\ref{tab:adaptive-state} counts radius-state events.
Table~\ref{tab:nx-runtime} uses 360 published outputs per method (two
conditions, three runs each), and Table~\ref{tab:vpr-recall} reports
cross-flight retrieval with explicit view budgets.

\subsection{Field image quality and interface continuity}
\label{sec:stitching-comparison}

Ours gives the lowest geometric error in the far field, the near field, and
the held-out sites, and the best frozen-detector results. We compare it with
four controls: Fixed Feather (\SI{30}{\metre} projection, confidence
feathering, no seam search), Fixed Radius (same projection, dynamic seams,
two-band fusion), Global Radius (one radius per panorama), and Per-seam (one
radius per overlap, 72-pixel seam search). Ours adds the wider adaptive seam
search and the gated mesh; calibration, exposure correction, and ERP sampling
are shared.

Geometry is the error of mutual AKAZE matches in fixed bands of the unblended
overlaps, summarized as medians of per-frame median and P90 errors; seam RGB
and CIEDE2000 \cite{sharma2005ciede2000} measure seam colour difference. Line
continuity counts LSD structures that cross or continue across four fixed
seam bands of the final ERP (Supplementary Sec.~S4). Frozen YOLOv10-N runs on
eight rectilinear views at $45^{\circ}$ yaw steps, with angular AP,
precision, and recall at $10^{\circ}$ on a 33-frame reference subset.

\begin{table*}[t]
  \centering
\caption{Field-image quality and frozen-YOLOv10-N results.
  Geometry uses 60 far-field and 58 near-field shared valid frames;
  Med./P90 are frame-median summaries in pixels. S. RGB med. is the seam
  RGB difference. Detection uses 181 matched frames; angular AP/precision/recall
  use 33 reference frames at $10^{\circ}$. Cov. is coverage (\%); Conf. is
  mean confidence. Bold/underlined values mark best/second-best displayed
  results; ties share styling.}
  \label{tab:stitching-summary}
  \footnotesize
  \setlength{\tabcolsep}{1.2pt}
  \begin{tabular*}{0.98\textwidth}{@{\extracolsep{\fill}}lccccccccccc}
    \toprule
    & \multicolumn{3}{c}{Far-field quality} &
      \multicolumn{3}{c}{Near-field quality} &
      \multicolumn{5}{c}{Frozen-YOLO detection metrics} \\
    \cmidrule(lr){2-4}\cmidrule(lr){5-7}\cmidrule(lr){8-12}
    Method & Med. $\downarrow$ & P90 $\downarrow$ & S. RGB med. $\downarrow$ &
      Med. $\downarrow$ & P90 $\downarrow$ & S. RGB med. $\downarrow$ &
      Cov. $\uparrow$ & Conf. $\uparrow$ &
      \shortstack{A-AP@$10^{\circ}$\\(\%) $\uparrow$} &
      \shortstack{A-Prec.@$10^{\circ}$\\(\%) $\uparrow$} &
      \shortstack{A-Rec.@$10^{\circ}$\\(\%) $\uparrow$} \\
    \midrule
    Fixed Feather & 5.82 & 9.92 & \underline{10.50} &
      \underline{11.25} & 25.91 & 23.92 &
      \underline{47.0} & 0.790 & 67.1 & 88.1 & 67.3 \\
    Fixed Radius & 5.82 & 9.92 & \textbf{5.00} &
      \underline{11.25} & 25.91 & 13.08 &
      45.9 & 0.790 & 73.4 & 91.1 & 74.5 \\
    Global Radius & 4.83 & 9.42 & \textbf{5.00} &
      12.11 & 25.90 & 12.33 &
      45.9 & 0.787 & 71.2 & 87.0 & 72.7 \\
    Per-seam & \underline{4.02} & \underline{5.96} & \textbf{5.00} &
      11.27 & \underline{19.54} & \underline{12.25} &
      45.9 & \underline{0.801} & \underline{79.6} & \underline{93.8} & \underline{81.8} \\
    \textbf{Ours} & \textbf{3.80} & \textbf{5.88} & \textbf{5.00} &
      \textbf{1.86} & \textbf{10.60} & \textbf{8.50} &
      \textbf{49.7} & \textbf{0.809} & \textbf{88.0} & \textbf{96.1} & \textbf{89.1} \\
    \bottomrule
\end{tabular*}
\end{table*}

\begin{figure}[!t]
  \centering
  \includegraphics[width=\columnwidth]{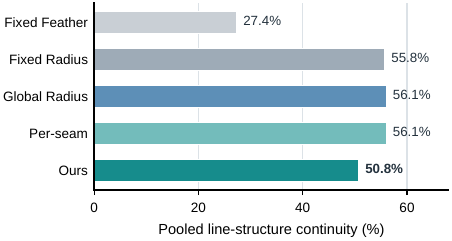}
\caption{Final-ERP line continuity on a separate continuous near-field
  sequence, computed as pooled preserved structures divided by pooled structure
  units across four fixed seam bands. Zero-structure seam--frame units do not
  contribute to either pooled count. Preserved structures per frame are 2.82,
  6.92, 6.75, 6.81, and 5.76 in the order plotted. Temporal-rule variants are compared
  in Table~\ref{tab:adaptive-state} and are not repeated here.}
  \label{fig:task-level-benefits}
\end{figure}

\begin{figure*}[t]
  \centering
  \includegraphics[width=\textwidth]{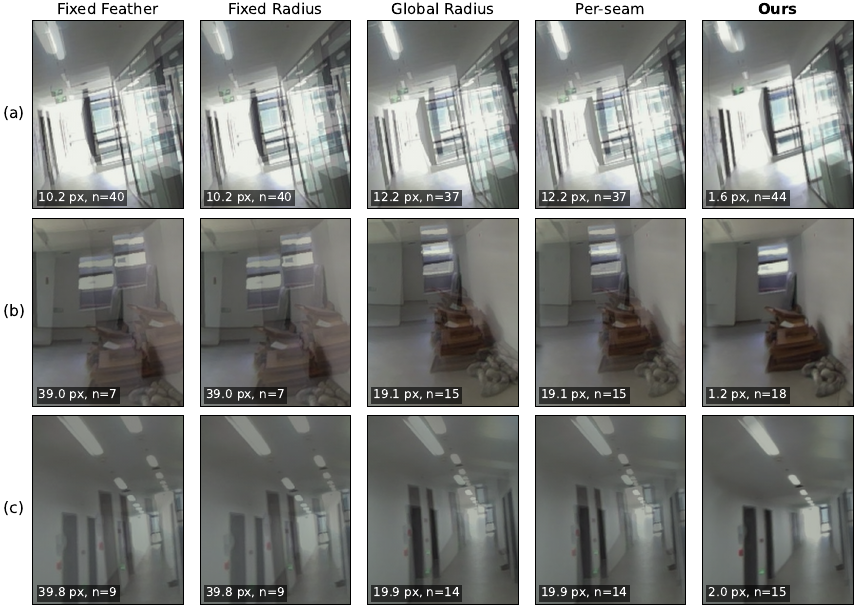}
  \caption{Near-field overlap bands for the five configurations, shown as a
  50/50 overlay of the two camera layers before seam selection and blending.
  Fixed Feather and Fixed Radius use the same \SI{30}{\metre} layers and
  differ only in blending, so their overlays are identical. In these overlaps
  Global Radius selects the same radius as Per-seam. Ours adds the gated mesh
  to Per-seam. The three examples were selected to show large misalignment and
  are not typical; Table~\ref{tab:stitching-summary} gives the median over all
  near-field frames. Numbers are AKAZE median errors and match counts in each
  band.}
  \label{fig:mesh-overlap}
\end{figure*}

In Table~\ref{tab:stitching-summary}, Ours reduces the error to 3.80/5.88 pixels far-field from 5.82/9.92 for Fixed Radius, and to 1.86/10.60 pixels
near-field from 11.27/19.54 for Per-seam, while cutting near-field seam RGB
from 12.25 to 8.50. It also gives the best detector confidence
(0.809) and angular AP/precision/recall (88.0/96.1/89.1\%).
Fig.~\ref{fig:mesh-overlap} shows why: in the two layers overlaid before
blending, the fixed \SI{30}{\metre} projection doubles near-field structure,
radius selection reduces it, and the gated mesh removes most of the remaining
offset.

The one exception is near-field line continuity
(Fig.~\ref{fig:task-level-benefits}), where Ours reaches 50.8\% against
56.1\% for Global Radius and Per-seam, 55.8\% for Fixed Radius, and 27.4\%
for Fixed Feather:
the mesh improves alignment, but continuity depends mainly on seam placement
and blending. Sec.~\ref{sec:nx-evaluation} finds the same in replay.

On the held-out Lake and South Square clips, with parameters fixed after
development, Ours again has the lowest errors, 5.23/8.95/9.98 pixels against
5.42/9.29/10.20 for Global Radius (Table~\ref{tab:cross-site-quality}), with
equal seam colour.

\begin{table}[t]
  \centering
\caption{Held-out Lake and South Square quality at $1280\times640$.
  Values are medians of frame-level metrics. Worst P90 is the largest
  overlap P90 per frame. Geometry is in pixels; seam RGB and CIEDE2000
  measure colour difference. Lower is better. Bold/underlining marks
  best/second-best values.}
  \label{tab:cross-site-quality}
  \footnotesize
  \setlength{\tabcolsep}{2.4pt}
  \begin{tabular*}{\columnwidth}{@{\extracolsep{\fill}}lccccc@{}}
    \toprule
    Method & Med. $\downarrow$ & P90 $\downarrow$ & \shortstack{Worst\\P90} $\downarrow$ &
      \shortstack{S. RGB\\med.} $\downarrow$ & \shortstack{CIEDE\\2000} $\downarrow$ \\
    \midrule
    Fixed Feather & 5.71 & 9.87 & 11.25 & \underline{9.33} & 4.073 \\
    Fixed Radius & 5.71 & 9.76 & 11.12 & \textbf{3.33} & 2.409 \\
    Global Radius & \underline{5.42} & \underline{9.29} & \underline{10.20} &
      \textbf{3.33} & \underline{2.406} \\
    Ours & \textbf{5.23} &
      \textbf{8.95} & \textbf{9.98} & \textbf{3.33} & \textbf{2.403} \\
    \bottomrule
  \end{tabular*}
\end{table}

\subsection{Temporal-policy ablation}

This ablation changes only the temporal acceptance rule. Two-confirmation
accepts a new radius after two observations, Immediate after one, and Ours
after one if the improvement is at least 0.5 pixels; the photometric fallback
always needs two. An Ours transition is accelerated when it agrees with
Immediate while Two-confirmation still differs, and confirmed when it agrees
with Two-confirmation. Table~\ref{tab:adaptive-state} counts these events at
retained update opportunities, where one opportunity is one overlap at one
retained evaluation frame. A switch is counted when an overlap's accepted
radius differs from that at the previous retained frame; the first frame has
no switch, and counts are pooled over the four overlaps.

\begin{table}[t]
  \centering
\caption{Temporal-policy ablation at $1280\times640$.
  Sw.: accepted radius-state transitions; Supp.: reduction relative to Immediate;
  Acc./Conf.: accelerated/confirmed events; Acc.: accelerated fraction;
  $\Delta$Line I/T: continuity difference from Immediate/Two-confirmation
  in percentage points. A switch is one accepted state change for one overlap
  between consecutive retained evaluation frames.}
  \label{tab:adaptive-state}
  \footnotesize
  \setlength{\tabcolsep}{1.2pt}
  \begin{tabular*}{\columnwidth}{@{\extracolsep{\fill}}lccccccc@{}}
    \toprule
    & \multicolumn{2}{c}{Endpoint switches} &
      \multicolumn{5}{c}{Ours} \\
    \cmidrule(lr){2-3}\cmidrule(lr){4-8}
    Condition & \shortstack{Two\\sw.} & \shortstack{Imm.\\sw.} &
      \shortstack{Ours\\sw.} & \shortstack{Supp.\\(\%)} &
      \shortstack{Acc./\\Conf.} & \shortstack{Acc.\\(\%)} & \shortstack{$\Delta$Line\\I/T} \\
    \midrule
    Near-field & 6 & 49 & 15 & 69.4 & 10/5 & 66.7 & $+0.25/-0.99$ \\
    Far-field & 7 & 19 & 6 & 68.4 & 5/1 & 83.3 & $+0.12/-0.20$ \\
    \bottomrule
  \end{tabular*}
\end{table}

Compared with Immediate, Ours reduces accepted radius transitions from 49 to
15 in the near field and from 19 to 6 in the far field, reductions of 69.4\%
and 68.4\%. Of these Ours transitions, 10/15 and 5/6 are accelerated and the
rest are confirmed. Table~\ref{tab:adaptive-state} also gives the
line-continuity differences to Immediate and Two-confirmation.

\begin{table*}[t]
\centering
\caption{Embedded operating characteristics per condition under the controlled 20 Hz NVIDIA Jetson
Orin NX replay in 25 W mode. Ours is the full method (adaptive seam selection with the gated residual
mesh); Per-seam is not included. Each entry is the mean over three runs of 60 captured ERPs, and
continuity is mean $\pm$ standard deviation across runs. This spread measures run-to-run
repeatability of a deterministic replay, not scene sampling, so it bounds reproducibility and does
not support a significance claim. All configurations keep the 20 Hz output rate with at most
0.1\% drop and stay within the 16 GB memory. Per-output continuity pools preserved and structure
counts over four fixed seam bands; outputs with zero support are omitted. Energy per output equals
power divided by output rate and is not listed. Bold marks a continuity mean separated from all
others in its condition by more than the run-to-run spread; none is bold in the indoor corridor.
Cost columns are not marked.}
\label{tab:nx-runtime}
\footnotesize
\setlength{\tabcolsep}{4pt}
\begin{tabular*}{\textwidth}{@{\extracolsep{\fill}}lcccccc@{}}
\toprule
& Output quality & \multicolumn{3}{c}{Real-time delivery} & \multicolumn{2}{c}{Embedded resources} \\
\cmidrule(lr){2-2}\cmidrule(lr){3-5}\cmidrule(l){6-7}
Method & \shortstack{Line continuity (\%)} $\uparrow$ & \shortstack{Output\\rate (Hz)} & \shortstack{Median\\completion (ms)} & \shortstack{Drop\\(\%)} & \shortstack{Power\\(W)} & \shortstack{Peak\\RAM (GiB)} \\
\midrule
\multicolumn{7}{@{}l}{\emph{Outdoor air}}\\
Fixed Feather & 39.24 $\pm$ 0.00 & 20.00 & 20.7 & 0.00 & 10.90 & 7.43 \\
Fixed Radius & 40.06 $\pm$ 0.00 & 20.00 & 20.6 & 0.00 & 10.89 & 7.39 \\
Global Radius & 55.19 $\pm$ 0.24 & 20.00 & 22.0 & 0.00 & 12.14 & 7.93 \\
Ours & \textbf{57.47 $\pm$ 0.11} & 19.98 & 44.7 & 0.10 & 13.05 & 8.02 \\
\midrule
\multicolumn{7}{@{}l}{\emph{Indoor corridor}}\\
Fixed Feather & 28.70 $\pm$ 0.00 & 20.00 & 19.9 & 0.00 & 10.76 & 7.35 \\
Fixed Radius & 23.86 $\pm$ 0.00 & 20.00 & 20.0 & 0.00 & 10.81 & 7.18 \\
Global Radius & 53.30 $\pm$ 0.33 & 20.00 & 21.1 & 0.02 & 11.81 & 7.88 \\
Ours & 53.74 $\pm$ 0.92 & 19.99 & 39.9 & 0.03 & 12.78 & 8.02 \\
\bottomrule
\end{tabular*}
\end{table*}

\subsection{Controlled onboard replay on NVIDIA Jetson Orin NX}
\label{sec:nx-evaluation}

Recorded synchronized four-fisheye flight groups are replayed on an NVIDIA
Jetson Orin NX Developer Kit (16\,GB, CUDA 11.4, TensorRT 8.5.2.2, OpenCV
4.8.0) in \SI{25}{\watt} mode with locked clocks. In each run, INA3221
\texttt{VDD\_IN} logs module-input power and \texttt{tegrastats} logs memory
and temperature. Inputs arrive at \SI{20}{\hertz}; a run is accepted only if
the input rate is $20.0\pm0.1$\,Hz, the P95 absolute arrival error is at most
5\,ms, and no decode fails. This gives repeatable, method-matched timing,
throughput, drop, power, and energy measurements on identical flight inputs.

Fixed Feather, Fixed Radius, Global Radius, and Ours run under the embedded
protocol of Sec.~\ref{sec:embedded}. Ours is the full method of
Sec.~\ref{sec:stitching-comparison}, adaptive seam selection with the gated
residual mesh; Per-seam is not part of the embedded comparison. Each method
runs three times per condition in rotated order, and the first
\SI{10}{\second} are excluded. Timing is from four-view availability to the
ERP completion timestamp. From each run, 60 preselected published ERPs after
warm-up are scored with the fixed-band line-continuity metric, and the same
run gives latency, throughput, timeliness, drop, power, and energy.

Unlike Fig.~\ref{fig:task-level-benefits}, which pools counts over one
sequence, Table~\ref{tab:nx-runtime} reports the mean per-output continuity
for each condition separately.

\paragraph{Operating characteristics.}
Table~\ref{tab:nx-runtime} reports quality and cost per condition. All four
configurations keep the \SI{20}{\hertz} output rate with at most 0.1\% drop
and stay inside the 16 GB memory, so the comparison is between delivered
quality and the latency, power, and memory spent for it.

Outdoors the full method is separated from every control. Its line
continuity is 57.47\% against 55.19\% for Global Radius, with run-to-run
spreads of 0.11 and 0.24 points, and it is more than 17 points above the two
fixed-radius controls. In the indoor corridor the two are not separated:
53.74\% against 53.30\%, a difference smaller than the 0.92-point spread of
the higher value. The cost appears in both conditions: median latency roughly
doubles, from 22.0 to \SI{44.7}{\milli\second} outdoors and from 21.1 to
\SI{39.9}{\milli\second} indoors, module-input power rises by about
\SI{0.9}{\watt}, and peak memory by about 0.1\,GiB.

\paragraph{Threshold-qualified yield.}
We also form a composite that counts an output as usable only if both its
continuity and its latency pass fixed thresholds. For a captured ERP $k$ with
continuity $C_k$ and scheduled completion latency $L_k$,
\begin{equation}
 \begin{aligned}
 z_k&=\mathbf{1}\!\left[C_k\geq60\%,\;L_k\leq100\,\mathrm{ms}\right],\\
 U_m&=\frac{\sum_{k\in m}z_k}{\sum_{r\in m}60E_r},
 \end{aligned}
 \label{eq:task-qualified-energy}
\end{equation}
where $E_r=\bar P_r/f_r$ is the run-level energy per completed output. With a
60\% threshold and a \SI{100}{\milli\second} deadline, the full method gives
0.645 qualified ERP/J and Global Radius 0.617. A paired hierarchical bootstrap
over three run labels and circular five-frame blocks (20,000 replicates) gives
a 95\% interval of $[-0.093,0.140]$ ERP/J with 0.678 of replicates above
zero, and the ordering reverses between conditions. Supplementary Fig.~S1 shows that the
ranking depends on both the threshold and the deadline. We therefore report
this composite only as a sensitivity surface in the supplement and do not use
it to rank the configurations; Table~\ref{tab:nx-runtime} carries the
comparison.

For system design, these results support a conditional use of the
additional formation stages. The continuity gain is clear outdoors and absent
in the corridor, while the latency and power cost is paid in both. The
operating environment decides whether the extra processing is worth it, and
no single configuration is best in all conditions.

\subsection{Frozen-model interface evaluations}
\label{sec:vpr-evaluation}

Visual place recognition (VPR) compares raw fisheye, rectified 4V, ERP 4V,
and ERP 8V inputs with a frozen ResNet-18 CosPlace model
\cite{Berton_CVPR_2022_CosPlace}. Inputs are preprocessed to $320\times320$,
and retrieval uses cosine similarity without temporal smoothing. Rectified 4V
and ERP 4V have the same FOV, raw fisheye 4V is a wider-FOV reference, and
the 4V and 8V settings need four and eight descriptor calls. Teach--repeat
queries are sampled every two seconds, and a retrieval is correct within a
\SI{10}{\metre} GNSS distance. Tests cover day, day-to-night, and
night-to-night conditions.

\begin{table}[t]
  \centering
\caption{Frozen-CosPlace cross-flight retrieval. Day R@1/R@5 average across
  daytime sites; Err. is the mean site-median top-1 GNSS error.
  Square D$\to$N and Playground N$\to$N report R@5 (\%).
  Calls are descriptor evaluations per group. Rectified 4V and ERP 4V
  use equal FOV. Bold/underlining marks best/second-best values.}
  \label{tab:vpr-recall}
  \footnotesize
  \setlength{\tabcolsep}{1.2pt}
    \begin{tabular*}{\columnwidth}{@{\extracolsep{\fill}}lcccccc@{}}
      \toprule
      Representation & Calls $\downarrow$ & \shortstack{Day\\R@1} $\uparrow$ &
        \shortstack{Day\\R@5} $\uparrow$ & \shortstack{Err.\\(m)} $\downarrow$ &
        \shortstack{Square\\D$\to$N} $\uparrow$ & \shortstack{Play.\\N$\to$N} $\uparrow$ \\
      \midrule
      Raw fisheye 4V & 4 & \underline{69.3} & 85.8 &
        \underline{18.1} & 9.5 & \underline{70.4} \\
      Rectified multi-view 4V & 4 & 61.8 & \underline{87.2} & 19.8 &
        \underline{19.0} & \underline{70.4} \\
      ERP 4V & 4 & 66.7 & 83.6 & 20.1 & \underline{19.0} &
        \textbf{77.8} \\
      ERP 8V & 8 & \textbf{74.9} & \textbf{90.8} & \textbf{17.2} &
        \textbf{23.8} & 63.0 \\
      \bottomrule
    \end{tabular*}
\end{table}

In the equal-FOV four-call comparison (Table~\ref{tab:vpr-recall}), ERP 4V
and rectified 4V give daytime R@1 of 66.7\% and 61.8\%, R@5 of 83.6\% and
87.2\%, and mean site-median GNSS errors of 20.1 and 19.8 m. With eight
calls, ERP 8V gives the highest daytime R@1/R@5 of 74.9\%/90.8\% and the
highest South Square day-to-night R@5 of 23.8\%, while ERP 4V gives the
highest Playground night-to-night R@5 of 77.8\%. These results document use of
the exported interfaces across view budgets and lighting conditions.

\subsection{External stitching baselines}
\label{sec:external-baselines}

The controls in Secs.~\ref{sec:stitching-comparison}--\ref{sec:nx-evaluation}
isolate stages inside one calibrated pipeline. Here we compare against
stitching software developed outside this work. All external methods run on
the outdoor recorded set with unlimited time, full resolution, and their
published default parameters, and receive the same calibrated ERP layers as
the onboard pipeline, so only alignment and composition differ. We supply the
layers because registration from image content alone is not reliable here: a
classical toolchain \cite{hugin2023} on the raw views linked all four cameras
in only 165 or 166 of the 181 frames. Seam selection is swept from none,
through a dynamic-programming seam, to a graph-cut seam
\cite{kwatra2003graphcut,boykov2004maxflow}, with multi-resolution blending
\cite{burt1983multiresolution} or a plain feather. External methods are not
paced at \SI{20}{\hertz}, since at seconds per frame a paced run would only
measure the drop rate; their throughput is the achieved rate.

\begin{table*}[t]
\centering
\caption{External stitching baselines on the recorded field-flight set. All
external configurations receive the same fixed calibrated spherical ERP layer
as the onboard pipeline, so only seam selection and composition differ; seam
and blending operators are OpenCV implementations \cite{bradski2000opencv},
and multi-band composition follows \cite{burt1983multiresolution}. Line
continuity and seam RGB use one implementation over identical fixed bands, so
they are comparable across the rows of this table but not with
Table~\ref{tab:stitching-summary} or the replay outputs of
Table~\ref{tab:nx-runtime}. Bold marks the best measured value in each
column.}
\label{tab:external-baselines}
\footnotesize
\setlength{\tabcolsep}{4pt}
\begin{tabular*}{\textwidth}{@{\extracolsep{\fill}}lccccccc@{}}
\toprule
& & \multicolumn{2}{c}{Output quality} & \multicolumn{3}{c}{Cost per completed output} \\
\cmidrule(lr){3-4}\cmidrule(l){5-7}
Method & \shortstack{Frames\\missed} & \shortstack{Line\\continuity (\%)} $\uparrow$ & \shortstack{Seam\\RGB} $\downarrow$ & \shortstack{Desktop\\latency (ms)} & \shortstack{NX\\latency (ms)\textsuperscript{a}} & \shortstack{NX energy\\(J/output)\textsuperscript{a}} \\
\midrule
Ours (onboard)\textsuperscript{b} & 0 & \textbf{58.82} & 2.00 & --- & \textbf{44.7} & \textbf{0.653} \\
Feather only\textsuperscript{c} & 0 & 57.14 & 1.67 & 72 & 263.7 & 2.51 \\
enblend seam ${+}$ multi-resolution blend \cite{hugin2023} & 0 & 58.33 & 1.67 & --- & 3466 & 31.6 \\
DP seam ${+}$ multi-band\textsuperscript{c} & 0 & 55.56 & \textbf{1.17} & 141 & 461.7 & 4.53 \\
Graph-cut seam \cite{kwatra2003graphcut,boykov2004maxflow} ${+}$ multi-band & 0 & 57.14 & \textbf{1.17} & 3479 & 7363 & 69.2 \\
APAP/Moving DLT \cite{zaragoza2013apap}\textsuperscript{d} & 15 & 57.69 & 1.83 & 2720 & --- & --- \\
Patch-alignment seam-cutting \cite{liao2025patchseam}\textsuperscript{e} & 0 & 53.85 & 2.17 & 16570 & --- & --- \\
\bottomrule
\end{tabular*}

\vspace{2pt}
{\footnotesize
\begin{flushleft}
\textsuperscript{a}\,Median of three repeat runs on the module in \SI{25}{\watt} mode;
module-input power from the INA3221 \texttt{VDD\_IN} channel. The frequency floor was not locked
for the enblend and graph-cut rows. A dash in the desktop column means the configuration was not
run on the desktop; the onboard pipeline has no desktop implementation.\\
\textsuperscript{b}\,Outdoor values of Table~\ref{tab:nx-runtime}: median latency from four-view
availability to ERP completion, and module-input power divided by the sustained output rate.\\
\textsuperscript{c}\,Latency and energy from the same three repeat runs per row in
\SI{25}{\watt} mode, with INA3221 \texttt{VDD\_IN} power sampled over each frame's processing
window. These are the only external configurations whose desktop cost is within a small factor
of the input period. A colour${+}$gradient seam cost changed continuity by at most 0.5 points
and is omitted.\\
\textsuperscript{d}\,No module figure: MATLAB with compiled extensions, and MATLAB is not
available for the module's architecture.\\
\textsuperscript{e}\,No module figure: our GNU Octave port could run on the module but was not
measured there, since at \SI{16.6}{\second} per frame on the desktop it is already 331 times
the input period. The published alignment stage is bypassed because the calibrated layers are
already registered, and composition uses the hard seam of the reference implementation.
\end{flushleft}}
\end{table*}

\textbf{Cost.} The onboard pipeline is the only configuration inside the
\SI{50}{\milli\second} input period: \SI{44.7}{\milli\second} and
\SI{0.653}{\joule} per output on the module. The two cheapest external
compositions, plain feather and a dynamic-programming seam, need 5.3 and 9.2
times the period and 3.8 and 6.9 times the energy per output. The external
configuration closest in continuity, enblend, costs 69 times the period, and
graph-cut costs 147 times the period and 106 times the energy. External
methods draw less power (8.5 to \SI{9.9}{\watt} against \SI{13.05}{\watt})
because they are CPU-bound, but their latency grows by a larger factor, so
energy per completed output is the relevant measure. On the desktop (Apple M2
Pro, CPU only) the two cheapest take 72 and \SI{141}{\milli\second}, which is
still above the period.

\textbf{Quality.} The onboard pipeline has the highest line continuity in
Table~\ref{tab:external-baselines}. A paired bootstrap over the 181 frames
(20\,000 replicates) puts it above every external configuration in mean
per-frame difference, by 1.86 to 5.89 points, with 95\% intervals excluding
zero. The per-frame median difference, a more conservative summary, separates
only the patch-alignment method, so we read quality as parity with external
composition, not superiority. The advantage is parity quality at a small
fraction of the cost.

\textbf{Alignment methods.} As-projective-as-possible warping with Moving DLT
\cite{zaragoza2013apap} is a strong aligner: its far-field unblended-overlap
feature error is \SI{0.63}{pixel}, against \SI{3.81}{pixel} for the full
method. This does not carry over to the final ERP: it misses 15 outdoor frames,
its four-overlap complete fit rate indoors is 13.3\%, its continuity is lower,
and it needs \SI{2.7}{\second} per frame. Learned and patch-based methods
estimate a warp between photographs of unknown relative geometry, which a
calibrated rig already knows, so we ran a method whose contribution is in
composition, patch alignment for seam-cutting \cite{liao2025patchseam}, with
its alignment bypassed. Its graph cut is seeded from the part of one image
lying outside the other; with $161^\circ$ layers both are valid across the
whole overlap and this seed is empty, so the adapter composites whole layers
pairwise around the ring. It completes every frame but needs
\SI{16.6}{\second} per frame, 331 times the period, and has the lowest
continuity and largest seam colour difference, consistent with its hard seam
and the stretched fisheye periphery that it does not suppress. This comes
from the transfer to a calibrated rig, not from a defect of the method. The
two alignment methods have no module figure (footnotes d and e); at 54 and
331 times the period on the desktop, a module run would not change the
conclusion.

\section{Conclusion}

We presented an open panoramic aerial robot with airframe-integrated
four-fisheye sensing, calibrated raw-view access, and an ERP interface.
Overlap-wise projection and gated local correction reduce geometric
misalignment on the evaluated samples, and seam-controlled fusion forms the
final ERP, which the onboard NVIDIA Jetson Orin NX computes from the live camera stream in
flight. Controlled NX replay measures final-ERP quality, output rate, and
module-input energy on common recorded inputs. Outdoors the full method gives
higher final-ERP line continuity than every control; in the indoor corridor
it matches the simpler radius control; in both cases it needs about twice the
latency and about one watt more power. Compared with external stitching
software given the same calibrated projection, the onboard configuration
gives final-ERP line continuity no lower than any tested method, while every
external configuration measured on the module needs 5.3 to 147 times the
input period and 3.8 to 106 times the energy per output. The formation
complexity should therefore be chosen for the operating conditions and the
application. Frozen detection and retrieval tests characterize the exported
views with explicit reference criteria and view budgets, and the released
CAD, calibration, software, and field data support reuse of the system.

\balance
\bibliographystyle{IEEEtran}
\bibliography{references}

@article{dai2026rflypano,
  author  = {Dai, Dun and Lu, Ze and Dai, Xunhua and Quan, Quan},
  title   = {{RflyPano}: A Panoramic Benchmark for Ultra-Low Altitude {UAV}
             Localization Powered by {RflySim}},
  journal = {Proceedings of the AAAI Conference on Artificial Intelligence},
  volume  = {40},
  number  = {22},
  pages   = {18216--18224},
  year    = {2026},
  doi     = {10.1609/aaai.v40i22.38884}
}

@misc{insta360x5_2025,
  author       = {{Insta360}},
  title        = {{X5} Flagship 360 Action Camera},
  year         = {2025},
  url          = {https://store.insta360.com/product/x5},
  note         = {Accessed 1 September 2026}
}

@misc{djiosmo360_2025,
  author       = {{DJI}},
  title        = {{Osmo 360} Specifications},
  year         = {2025},
  url          = {https://www.dji.com/360/specs},
  note         = {Accessed 1 September 2026}
}

@misc{micoair2026h743,
  author       = {{MicoAir Technology}},
  title        = {{MicoAir743v2-AIO-45A} User Manual},
  year         = {2026},
  url          = {https://micoair.cn/zh/docs/flight-controller/micoair743-aio-series/micoair743v2-aio-45a-manual},
  note         = {Accessed 26 July 2026}
}

@misc{micoair2026mgf10,
  author       = {{MicoAir Technology}},
  title        = {{MG-F10-A} Dual-Frequency {GNSS} Module User Manual},
  year         = {2026},
  note         = {Accessed 26 July 2026},
  url          = {https://micoair.cn/zh/docs/gps-rtk/mg-f10/mg-f10-a-gnss}
}

@inproceedings{furgale2013calibration,
  author    = {Furgale, Paul and Rehder, Joern and Siegwart, Roland},
  title     = {Unified Temporal and Spatial Calibration for Multi-Sensor
               Systems},
  booktitle = {2013 IEEE/RSJ International Conference on Intelligent Robots
               and Systems},
  pages     = {1280--1286},
  year      = {2013},
  doi       = {10.1109/IROS.2013.6696514}
}

@article{li2024marslvig,
  author  = {Li, Haotian and Zou, Yuying and Chen, Nan and Lin, Jiarong and
             Liu, Xiyuan and Xu, Wei and Zheng, Chunran and Li, Rundong and
             He, Dongjiao and Kong, Fanze and Cai, Yixi and Liu, Zheng and
             Zhou, Shunbo and Xue, Kaiwen and Zhang, Fu},
  title   = {{MARS-LVIG} Dataset: A Multi-Sensor Aerial Robots {SLAM} Dataset
             for {LiDAR}-Visual-Inertial-{GNSS} Fusion},
  journal = {The International Journal of Robotics Research},
  volume  = {43},
  number  = {8},
  pages   = {1114--1127},
  year    = {2024},
  doi     = {10.1177/02783649241227968}
}

@inproceedings{wang2025uavscenes,
  author    = {Wang, Sijie and Li, Siqi and Zhang, Yawei and Yu, Shangshu and
               Yuan, Shenghai and She, Rui and Guo, Quanjiang and Zheng, Jinxuan
               and Howe, Ong Kang and Chandra, Leonrich and Srijeyan, Shrivarshann
               and Sivadas, Aditya and Aggarwal, Toshan and Liu, Heyuan and
               Zhang, Hongming and Chen, Chujie and Jiang, Junyu and Xie, Lihua
               and Tay, Wee Peng},
  title     = {{UAVScenes}: A Multi-Modal Dataset for {UAVs}},
  booktitle = {Proceedings of the IEEE/CVF International Conference on Computer
               Vision},
  pages     = {28946--28958},
  year      = {2025}
}

@article{wang2022aerialpass,
  author  = {Wang, Jia and Yang, Kailun and Gao, Shaohua and Sun, Lei and
             Zhu, Chengxi and Wang, Kaiwei and Bai, Jian},
  title   = {High-Performance Panoramic Annular Lens Design for Real-Time
             Semantic Segmentation on Aerial Imagery},
  journal = {Optical Engineering},
  volume  = {61},
  number  = {3},
  pages   = {035101},
  year    = {2022},
  doi     = {10.1117/1.OE.61.3.035101}
}

@inproceedings{huang2024_360loc,
  author    = {Huang, Huajian and Liu, Changkun and Zhu, Yipeng and Cheng, Hui
               and Braud, Tristan and Yeung, Sai-Kit},
  title     = {{360Loc}: A Dataset and Benchmark for Omnidirectional Visual
               Localization with Cross-Device Queries},
  booktitle = {Proceedings of the IEEE/CVF Conference on Computer Vision and
               Pattern Recognition},
  pages     = {22314--22324},
  year      = {2024}
}

@inproceedings{gochoo2023fisheye8k,
  author    = {Gochoo, Munkhjargal and Otgonbold, Munkh-Erdene and Ganbold,
               Erkhembayar and Hsieh, Jun-Wei and Chang, Ming-Ching and Chen,
               Ping-Yang and Dorj, Byambaa and Al Jassmi, Hamad and Batnasan,
               Ganzorig and Alnajjar, Fady and Abduljabbar, Mohammed and Lin,
               Fang-Pang},
  title     = {{FishEye8K}: A Benchmark and Dataset for Fisheye Camera Object
               Detection},
  booktitle = {Proceedings of the IEEE/CVF Conference on Computer Vision and
               Pattern Recognition Workshops},
  pages     = {5305--5313},
  year      = {2023}
}

@inproceedings{zhang2022trans4pass,
  author    = {Zhang, Jiaming and Yang, Kailun and Ma, Chaoxiang and
               Rei{\ss}, Simon and Peng, Kunyu and Stiefelhagen, Rainer},
  title     = {Bending Reality: Distortion-Aware Transformers for Adapting to
               Panoramic Semantic Segmentation},
  booktitle = {Proceedings of the IEEE/CVF Conference on Computer Vision and
               Pattern Recognition},
  pages     = {16917--16927},
  year      = {2022}
}

@inproceedings{zhang2024goodsam,
  author    = {Zhang, Weiming and Liu, Yexin and Zheng, Xu and Wang, Lin},
  title     = {{GoodSAM}: Bridging Domain and Capacity Gaps via Segment Anything
               Model for Distortion-Aware Panoramic Semantic Segmentation},
  booktitle = {Proceedings of the IEEE/CVF Conference on Computer Vision and
               Pattern Recognition},
  pages     = {28264--28273},
  year      = {2024}
}

@inproceedings{nie2023parallaxstitching,
  author    = {Nie, Lang and Lin, Chunyu and Liao, Kang and Liu, Shuaicheng and
               Zhao, Yao},
  title     = {Parallax-Tolerant Unsupervised Deep Image Stitching},
  booktitle = {Proceedings of the IEEE/CVF International Conference on Computer
               Vision},
  pages     = {7399--7408},
  year      = {2023}
}

@inproceedings{zaragoza2013apap,
  author    = {Zaragoza, Julio and Chin, Tat-Jun and Brown, Michael S. and
               Suter, David},
  title     = {As-Projective-As-Possible Image Stitching with Moving {DLT}},
  booktitle = {Proceedings of the IEEE Conference on Computer Vision and
               Pattern Recognition},
  pages     = {2339--2346},
  year      = {2013},
  doi       = {10.1109/CVPR.2013.303}
}

@inproceedings{lin2015aanap,
  author    = {Lin, Chung-Ching and Pankanti, Sharathchandra U. and
               Natesan Ramamurthy, Karthikeyan and Aravkin, Aleksandr Y.},
  title     = {Adaptive As-Natural-As-Possible Image Stitching},
  booktitle = {Proceedings of the IEEE Conference on Computer Vision and
               Pattern Recognition},
  pages     = {1155--1163},
  year      = {2015}
}

@inproceedings{du2022gesgsp,
  author    = {Du, Peng and Ning, Jifeng and Cui, Jiguang and Huang, Shaoli and
               Wang, Xinchao and Wang, Jiaxin},
  title     = {Geometric Structure Preserving Warp for Natural Image
               Stitching},
  booktitle = {Proceedings of the IEEE/CVF Conference on Computer Vision and
               Pattern Recognition},
  pages     = {3688--3696},
  year      = {2022}
}

@inproceedings{ding2021gravitystitching,
  author    = {Ding, Yaqing and Barath, Daniel and Kukelova, Zuzana},
  title     = {Minimal Solutions for Panoramic Stitching Given Gravity Prior},
  booktitle = {Proceedings of the IEEE/CVF International Conference on Computer
               Vision},
  pages     = {5579--5588},
  year      = {2021}
}

@inproceedings{alibey2023mixvpr,
  author    = {Ali-bey, Amar and Chaib-draa, Brahim and Gigu{\`e}re, Philippe},
  title     = {{MixVPR}: Feature Mixing for Visual Place Recognition},
  booktitle = {Proceedings of the IEEE/CVF Winter Conference on Applications of
               Computer Vision},
  pages     = {2998--3007},
  year      = {2023}
}

@inproceedings{lindenberger2023lightglue,
  author    = {Lindenberger, Philipp and Sarlin, Paul-Edouard and Pollefeys,
               Marc},
  title     = {{LightGlue}: Local Feature Matching at Light Speed},
  booktitle = {Proceedings of the IEEE/CVF International Conference on Computer
               Vision},
  pages     = {17627--17638},
  year      = {2023}
}

@misc{wu2026panoair,
  author        = {Wu, Yiyang and Zhang, Xiaohu and Du, Yanjin and Zhang,
                   Tongsu and Li, Chujun and Chen, Siyang and Zhang, Guoyi and
                   Xu, Xiangpeng},
  title         = {{PanoAir}: A Panoramic Visual-Inertial {SLAM} with
                   Cross-Time Real-World {UAV} Dataset},
  year          = {2026},
  eprint        = {2604.00852},
  archiveprefix = {arXiv},
  primaryclass  = {cs.RO},
  doi           = {10.48550/arXiv.2604.00852},
  url           = {https://arxiv.org/abs/2604.00852}
}

@inproceedings{ge2026airsim360,
  author    = {Ge, Xian and Pan, Yuling and Zhang, Yuhang and Li, Xiang and
               Zhang, Weijun and Zhang, Dizhe and Wan, Zhaoliang and Lin, Xin
               and Zhang, Xiangkai and Liang, Juntao and Li, Xiangtai and Jiang,
               WenJie and Du, Bo and Yang, Ming-Hsuan and Qi, Lu},
  title     = {{AirSim360}: A Panoramic Simulation Platform within Drone View},
  booktitle = {Proceedings of the IEEE/CVF Conference on Computer Vision and
               Pattern Recognition},
  pages     = {26931--26940},
  year      = {2026}
}

@article{kim2024pair360,
  author  = {Kim, Geunu and Kim, Daeho and Jang, Jaeyun and Hwang, Hyoseok},
  title   = {{PAIR360}: A Paired Dataset of High-Resolution $360^{\circ}$
             Panoramic Images and {LiDAR} Scans},
  journal = {IEEE Robotics and Automation Letters},
  volume  = {9},
  number  = {11},
  pages   = {9550--9557},
  year    = {2024},
  doi     = {10.1109/LRA.2024.3460418}
}

@inproceedings{liao2025patchseam,
  author    = {Liao, Tianli and Zhao, Chenyang and Li, Lei and Cao, Heling},
  title     = {Leveraging Local Patch Alignment to Seam-Cutting for Large
               Parallax Image Stitching},
  booktitle = {Proceedings of the IEEE/CVF International Conference on Computer
               Vision},
  pages     = {27262--27271},
  year      = {2025}
}

@inproceedings{benny2025sphereuformer,
  author    = {Benny, Yaniv and Wolf, Lior},
  title     = {{SphereUFormer}: A U-Shaped Transformer for Spherical
               $360^{\circ}$ Perception},
  booktitle = {Proceedings of the IEEE/CVF Conference on Computer Vision and
               Pattern Recognition},
  pages     = {940--950},
  year      = {2025}
}

@inproceedings{cao2025panda,
  author    = {Cao, Zidong and Zhu, Jinjing and Zhang, Weiming and Ai, Hao and
               Bai, Haotian and Zhao, Hengshuang and Wang, Lin},
  title     = {{PanDA}: Towards Panoramic Depth Anything with Unlabeled
               Panoramas and M{\"o}bius Spatial Augmentation},
  booktitle = {Proceedings of the IEEE/CVF Conference on Computer Vision and
               Pattern Recognition},
  pages     = {982--992},
  year      = {2025}
}

@article{xu2022omniswarm,
  author  = {Xu, Hao and Zhang, Yichen and Zhou, Boyu and Wang, Luqi and Yao,
             Xinjie and Meng, Guotao and Shen, Shaojie},
  title   = {{Omni-Swarm}: A Decentralized Omnidirectional
             Visual--Inertial--{UWB} State Estimation System for Aerial Swarms},
  journal = {IEEE Transactions on Robotics},
  volume  = {38},
  number  = {6},
  pages   = {3374--3394},
  year    = {2022},
  doi     = {10.1109/TRO.2022.3182503}
}

@inproceedings{Berton_CVPR_2022_CosPlace,
  author    = {Berton, Gabriele and Masone, Carlo and Caputo, Barbara},
  title     = {Rethinking Visual Geo-Localization for Large-Scale Applications},
  booktitle = {Proceedings of the IEEE/CVF Conference on Computer Vision and
               Pattern Recognition},
  pages     = {4878--4888},
  year      = {2022}
}

@inproceedings{kweon2023pixelwarping,
  author    = {Kweon, Hyeokjun and Kim, Hyeonseong and Kang, Yoonsu and Yoon,
               Youngho and Jeong, Wooseong and Yoon, Kuk-Jin},
  title     = {Pixel-Wise Warping for Deep Image Stitching},
  booktitle = {Proceedings of the AAAI Conference on Artificial Intelligence},
  volume    = {37},
  number    = {1},
  pages     = {1196--1204},
  year      = {2023},
  doi       = {10.1609/aaai.v37i1.25202}
}

@inproceedings{li2023sgat4pass,
  author    = {Li, Xuewei and Wu, Tao and Qi, Zhongang and Wang, Gaoang and
               Shan, Ying and Li, Xi},
  title     = {{SGAT4PASS}: Spherical Geometry-Aware Transformer for Panoramic
               Semantic Segmentation},
  booktitle = {Proceedings of the Thirty-Second International Joint Conference
               on Artificial Intelligence},
  pages     = {1125--1133},
  year      = {2023},
  doi       = {10.24963/ijcai.2023/125}
}

@inproceedings{reyarea2022_360monodepth,
  author    = {Rey-Area, Manuel and Yuan, Mingze and Richardt, Christian},
  title     = {{360MonoDepth}: High-Resolution $360^{\circ}$ Monocular Depth
               Estimation},
  booktitle = {Proceedings of the IEEE/CVF Conference on Computer Vision and
               Pattern Recognition},
  pages     = {3762--3772},
  year      = {2022},
  doi       = {10.1109/CVPR52688.2022.00374}
}

@article{keetha2024anyloc,
  author  = {Keetha, Nikhil and Mishra, Avneesh and Karhade, Jay and
             Jatavallabhula, Krishna Murthy and Scherer, Sebastian and Krishna,
             Madhava and Garg, Sourav},
  title   = {{AnyLoc}: Towards Universal Visual Place Recognition},
  journal = {IEEE Robotics and Automation Letters},
  volume  = {9},
  number  = {2},
  pages   = {1286--1293},
  year    = {2024},
  doi     = {10.1109/LRA.2023.3343602}
}

@article{sharma2005ciede2000,
  author  = {Sharma, Gaurav and Wu, Wencheng and Dalal, Edul N.},
  title   = {The {CIEDE2000} Color-Difference Formula: Implementation Notes,
             Supplementary Test Data, and Mathematical Observations},
  journal = {Color Research \& Application},
  volume  = {30},
  number  = {1},
  pages   = {21--30},
  year    = {2005},
  doi     = {10.1002/col.20070}
}

@inproceedings{liu2024omninxt,
  author    = {Liu, Peize and Feng, Chen and Xu, Yang and Ning, Yan and Xu, Hao
               and Shen, Shaojie},
  title     = {{OmniNxt}: A Fully Open-Source and Compact Aerial Robot With
               Omnidirectional Visual Perception},
  booktitle = {2024 IEEE/RSJ International Conference on Intelligent Robots and
               Systems (IROS)},
  pages     = {10605--10612},
  year      = {2024},
  doi       = {10.1109/IROS58592.2024.10802134}
}

@misc{omninxt2026hardware,
  author       = {{HKUST Aerial Robotics Group}},
  title        = {{OmniNxt}: Official Hardware and Software Release},
  howpublished = {GitHub repository},
  url          = {https://github.com/HKUST-Aerial-Robotics/OmniNxt},
  note         = {Accessed: Sep. 17, 2026}
}

@article{petrlik2025subterranean,
  author  = {Petrl{\'i}k, Mat{\v e}j and Petr{\'a}{\v c}ek, Pavel and
             Kr{\'a}tk{\'y}, V{\'i}t and Musil, Tom{\'a}{\v s} and
             Stasinchuk, Yurii and Vrba, Matou{\v s} and B{\'a}{\v c}a,
             Tom{\'a}{\v s} and He{\v r}t, Daniel and Pecka, Martin and
             Svoboda, Tom{\'a}{\v s} and Saska, Martin},
  title   = {{UAVs} Beneath the Surface: Cooperative Autonomy for
             Subterranean Search and Rescue in {DARPA SubT}},
  journal = {IEEE Transactions on Field Robotics},
  volume  = {2},
  pages   = {643--689},
  year    = {2025},
  doi     = {10.1109/TFR.2024.3492160}
}

@article{mattamala2025forest,
  author  = {Mattamala, Mat{\'i}as and Chebrolu, Nived and Frey, Jonas and
             Frei{\ss}muth, Leonard and Oh, Haedam and Casseau, Benoit and
             Hutter, Marco and Fallon, Maurice},
  title   = {Building Forest Inventories With Autonomous Legged Robots---System,
             Lessons, and Challenges Ahead},
  journal = {IEEE Transactions on Field Robotics},
  volume  = {2},
  pages   = {418--436},
  year    = {2025},
  doi     = {10.1109/TFR.2025.3583972}
}

@article{lin2018aerialnavigation,
  author  = {Lin, Yi and Gao, Fei and Qin, Tong and Gao, Wenliang and Liu,
             Tianbo and Wu, William and Yang, Zhenfei and Shen, Shaojie},
  title   = {Autonomous Aerial Navigation Using Monocular Visual--Inertial
             Fusion},
  journal = {Journal of Field Robotics},
  volume  = {35},
  number  = {1},
  pages   = {23--51},
  year    = {2018},
  doi     = {10.1002/rob.21732}
}

@article{grelsson2020horizonnet,
  author  = {Grelsson, Bertil and Robinson, Andreas and Felsberg, Michael and
             Khan, Fahad Shahbaz},
  title   = {{GPS}-Level Accurate Camera Localization With {HorizonNet}},
  journal = {Journal of Field Robotics},
  volume  = {37},
  number  = {6},
  pages   = {951--971},
  year    = {2020},
  doi     = {10.1002/rob.21929}
}

@article{lowe2004sift,
  author  = {Lowe, David G.},
  title   = {Distinctive Image Features From Scale-Invariant Keypoints},
  journal = {International Journal of Computer Vision},
  volume  = {60},
  number  = {2},
  pages   = {91--110},
  year    = {2004},
  doi     = {10.1023/B:VISI.0000029664.99615.94}
}

@inproceedings{kwatra2003graphcut,
  author    = {Kwatra, Vivek and Sch{\"o}dl, Arno and Essa, Irfan and Turk, Greg and Bobick, Aaron},
  title     = {Graphcut Textures: Image and Video Synthesis Using Graph Cuts},
  booktitle = {ACM SIGGRAPH 2003 Papers},
  pages     = {277--286},
  year      = {2003}
}

@article{boykov2004maxflow,
  author  = {Boykov, Yuri and Kolmogorov, Vladimir},
  title   = {An Experimental Comparison of Min-Cut/Max-Flow Algorithms for Energy Minimization in Vision},
  journal = {IEEE Transactions on Pattern Analysis and Machine Intelligence},
  volume  = {26},
  number  = {9},
  pages   = {1124--1137},
  year    = {2004}
}

@article{burt1983multiresolution,
  author  = {Burt, Peter J. and Adelson, Edward H.},
  title   = {A Multiresolution Spline with Application to Image Mosaics},
  journal = {ACM Transactions on Graphics},
  volume  = {2},
  number  = {4},
  pages   = {217--236},
  year    = {1983}
}

@article{bradski2000opencv,
  author  = {Bradski, Gary},
  title   = {The {OpenCV} Library},
  journal = {Dr. Dobb's Journal of Software Tools},
  volume  = {25},
  number  = {11},
  pages   = {120--125},
  year    = {2000}
}

@misc{hugin2023,
  author       = {{Hugin Project}},
  title        = {Hugin: Panorama photo stitcher},
  howpublished = {Version 2023.0, panotools},
  year         = {2023},
  note         = {[Online]. Available: \url{https://hugin.sourceforge.io}}
}

\end{document}